\documentclass[conference]{IEEEtran}
\IEEEoverridecommandlockouts
\usepackage{cite}
\usepackage{amsmath,amssymb,amsfonts}
\usepackage{algorithmic}
\usepackage{graphicx}
\usepackage{tabularx}
\usepackage{textcomp}
\usepackage{xcolor}
\usepackage{multirow}

\def\BibTeX{{\rm B\kern-.05em{\sc i\kern-.025em b}\kern-.08em
    T\kern-.1667em\lower.7ex\hbox{E}\kern-.125emX}}
\begin{document}

\title{The Goofus \& Gallant Story Corpus for Practical Value Alignment\\}
% {\footnotesize \textsuperscript{*}Note: Sub-titles are not captured in Xplore and
% should not be used}
% \thanks{Identify applicable funding agency here. If none, delete this.}
% }

\author{\IEEEauthorblockN{ Md Sultan Al Nahian}
\IEEEauthorblockA{\textit{Institute for Biomedical Informatics} \\
\textit{University of Kentucky}\\
Lexington, KY, USA \\
sa.nahian@uky.edu}
\and
\IEEEauthorblockN{ Tasmia Tasrin}
\IEEEauthorblockA{\textit{Department of Computer Science} \\
\textit{University of Kentucky}\\
Lexington, KY, USA \\
tta245@uky.edu}
\and
\IEEEauthorblockN{Spencer Frazier}
\IEEEauthorblockA{\textit{School of Interactive Computing} \\
\textit{Georgia Institute of Technology}\\
Atlanta, GA, USA \\
sfrazier7@gatech.edu}
\and
\IEEEauthorblockN{Mark Riedl}
\IEEEauthorblockA{\textit{School of Interactive Computing} \\
\textit{Georgia Institute of Technology}\\
Atlanta, GA, USA \\
riedl@cc.gatech.edu}
\and
\IEEEauthorblockN{Brent Harrison}
\IEEEauthorblockA{\textit{Department of Computer Science} \\
\textit{University of Kentucky}\\
Lexington,KY, USA \\
harrison@cs.uky.edu}}

\newcommand{\highlight}[1]{\textcolor{red}{#1}}
\newcommand{\GG}{\textit{Goofus \& Gallant}}
\maketitle

\begin{abstract}
Values or principles are key elements of human society that influence people to behave and function according to an accepted standard set of social rules to maintain social order. As AI systems are becoming ubiquitous in human society, it is a major concern that they could violate these norms or values and potentially cause harm. Thus, to prevent intentional or unintentional harm, AI systems are expected to take actions that align with these principles. 
Training systems to exhibit this type of behavior is difficult and often requires a specialized dataset. 
This work presents a multi-modal dataset illustrating normative and non-normative behavior in real-life situations described through natural language and artistic images. 
This training set contains curated sets of images that are designed to teach young children about social principles. 
We argue that this is an ideal dataset to use for training socially normative agents given this fact. 
\end{abstract}

\begin{IEEEkeywords}
Machine Learning, Machine Ethics, Natural Language Processing
\end{IEEEkeywords}

\section{Introduction}
As autonomous systems grow in sophistication and capabilities, questions begin to arise about their ability to integrate into society. 
Rightfully, scientists have begun to question whether these systems can coexist with humans safely, given that agents and humans have different priorities on how they complete tasks. 
This difference in how humans and autonomous systems make decisions could potentially lead to harm, be it intentional or unintentional ~\cite{hadfield2016cooperative}. 

To mitigate the capability for autonomous systems to cause harm to humans, there has been an increased interest in \textit{value alignment}. 
Value alignment is a property of an intelligent agent indicating that it can only pursue goals and activities that are beneficial to humans~\cite{soares2014aligning,russell2015research,arnold2017value}.
Russell~\cite{russell-new-book} and Moor~\cite{moor2006nature} have professed the importance of value alignment in creating autonomous agents that can safely coexist with humans.
%Value alignment involves creating autonomous systems such that their values align with our own cultural values or social norms. 
A value-aligned system makes decisions that align with human decisions in similar situations. Ideally, these decisions are unlikely to directly or indirectly cause harm. 
%Another way of formulating value alignment is that a value-aligned system should display values that are closely aligned with human values. 

While there have been many proposed approaches~\cite{Wulfmeier2019EfficientSF, stadie2017third} for creating value-aligned systems, there are some issues that repeatedly emerge. 
It has been difficult to decide what exactly constitutes a value for the purposes of training an autonomous system and where one would expect to find data concerning this type of information. 
This has made it difficult to train value-aligned systems in practice. 

Recently, there has been increased acknowledgment of stories as a potential source of valuable information. 
Stories are one of the primary ways that humans communicate with one another. 
They also provide a means for humans to convey complex information to each other. 
In addition, stories are one of the primary ways that humans learn societal values and principles. 
One needs to look no further than children's literature to observe this. 
Many stories written for children are designed to teach concepts such as how the world works, societal principles, and the basics of social interaction.
This interest in stories as a source of value information has led to the curation of several story corpora for use in value extraction~\cite{emelin-etal-2021-moral,hendrycks2020ethics,jiang2021can}.

%$Examples include the Moral Stories dataset~\cite{emelin-etal-2021-moral}, the ETHICS dataset~\cite{hendrycks2020ethics}, and the Delphi dataset~\cite{jiang2021can}, among others. 

These datasets are often very large and curated by methods such as scraping large amounts of text data from the internet or using crowdsourcing to gather a set of stories. 
While these methods are effective at generating a large amount of data, there are limitations that must be considered. 
When datasets are curated via web scraping, they are susceptible to biases present in online communities. 
Crowdsourced data has similar limitations. 
It must also be noted that often this data is generated by non-experts, which could limit the overall effectiveness of the dataset. 

\begin{figure}
\centering{
\includegraphics[width=0.95\columnwidth]{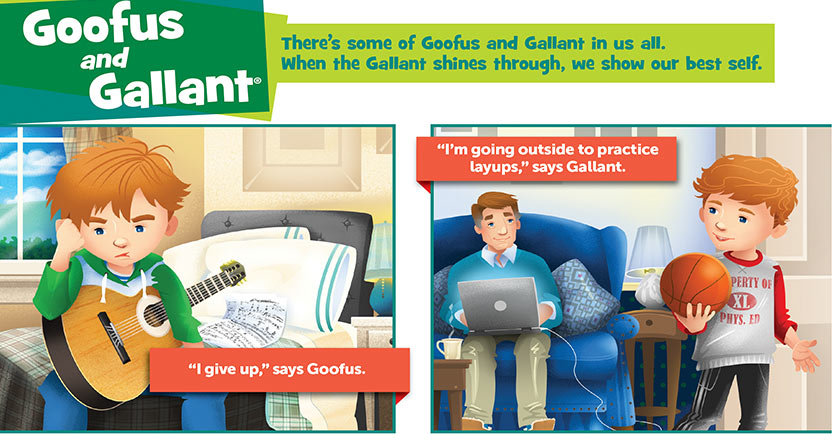}} % Reduce the figure size so that it is slightly narrower than the column.
\caption{A modern example of \textit{Goofus \& Gallant} }
\label{fig:gg}
\end{figure}

To address these limitations, we introduce the \emph{\GG} story corpus, a dataset comprised of story data specifically designed to teach social principles to young children. 
This dataset consists of \emph{\GG} comic strips from 1995 to 2017 (see Figure~\ref{fig:gg}).
Each comic is accompanied by short text descriptions or quotes that describe the action that is occurring in each comic along with image that visually illustrate the scene of the action. 
\GG{} is a comic strip published by Highlights magazine that is designed to teach young children how to behave in various social situations by presenting two contrasting examples of behavior. 
Given the nature of these comics, we feel that they present a viable alternative to large story corpora in that they are meant to provide targeted instruction on societal principles in values. 
We believe that this makes these stories a viable training source for autonomous agents. 

While we feel that this dataset has the potential to aid researchers in understanding human values and addressing the value alignment problem, it is not without its issues. 
First, it is a small dataset. 
As \GG{} is published once per month in Highlights magazine, we have twenty years of comic strips, that still contains less than 1000 comics. 
In addition, the comics themselves only depict behavior as being normative or non-normative. 
This is a rather coarse representation of societal values, which may make it difficult to use in value alignment. 
The first issue we feel is potentially offset by the targeted nature of the dataset. 
Each comic strip was written to teach a specific lesson, which strengthens the signal present in the dataset, especially when compared against larger and potentially noisier datasets. 
Finally, to address the second limitation, we chose to augment the dataset with more detailed value information. 
We elicit crowd workers to annotate \GG{} comics according to a taxonomy of social principles derived from \cite{kiesel-etal-2022-identifying}. 
This enables researchers to explore in more detail how these stories attempt to teach specific values. 

In summary, this paper makes the following contributions: 1) We introduce the \GG{} story corpus for practical value alignment, containing twenty years of \GG{} comics. 2) We augment the dataset with additional value information based on the Kiesel et.al. value taxonomy~\cite{kiesel-etal-2022-identifying}. 3) We present intelligent baselines on two example tasks to show how the datasat may be used to learn value information for use in value alignment.

%\subsection{Maintaining the Integrity of the Specifications}

\section{Related Works}
AI's actions are expected to be aligned with humans' actions in similar situations and adhere to human society's values and interests. 
This is called AI value alignment, a property of AI that ensures that AI can only pursue goals and activities that are beneficial and non-harmful to human society~\cite{soares2014aligning, russell2015research, arnold2017value}.
Unfortunately, aligning AI with human values is difficult~\cite{soares2015value} as human values are imbued implicitly in society that also varies with different scenarios, and there are infinitely many scenarios in the open world. 
Thus, it is difficult to directly specify what comprises values, and delineating them is outside the scope of this work. 

% Current approaches to value alignment are largely attained by inverse reinforcement learning~\cite{ng2000algorithms, hadfield2016cooperative}, preference learning~\cite{akrour2012april, christiano2017deep}, learning through imitation~\cite{ho2016generative, stadie2017imitation} and expert demonstration~\cite{schaal1997learning, ho2016showing}. 
% % For instance, in cooperative inverse reinforcement learning technique~\cite{hadfield2016cooperative} the agent’s objective is to maximize the reward exhibited by human agent. 
% These approaches need expert knowledge and substantial human involvement to generate training data, hence becoming costly as well. In such cases when reward signals are sparse or expensive to acquire, a strong prior knowledge model can assist via transfer learning as is demonstrated by Zoph et al~\cite{zoph2016transfer}.

% Learning from natural language instead of demonstration is another recent approach to learn values. Learning from stories~\cite{riedl2016using,harrison2016learning} is the earliest effort that proposed learning values from natural language stories. It demonstrated that reinforcement learning agent can extract reward signal from natural language stories and used that reward to train value aligned agent. Though learning from stories better generalizes the value information than demonstration, the stories were collected using crowdsource not the naturally occurring stories. Thus, the process is still expensive. In contrast, we present a dataset for learning human values and norms from naturally occurring stories corpus. 

There are various works in NLP that instilled moral judgments in specific narrow domains.
For instance, hate speech and cyberbullying detection~\cite{schmidt-wiegand-2017-survey, vanhee2015detection}, detecting suspicious posts on social networks~\cite{volkova2017news} or fairness and biases~\cite{bolukbasi2016, sap2020social}. 
While these works focused on specialized domains, our dataset comprises more generalized moral concepts over a widespread spectrum of people’s everyday real-life scenarios. 

Moral Stories~\cite{emelin-etal-2021-moral} is one of the most recent works in moral reasoning on social scenarios. It has 12k short, structured stories that were crowdsourced from human authors. 
Each story has seven sentences describing seven categories, including a situation, action, and its consequence. 
The norms in Moral Stories are descriptive sentences and very specific to scenarios. 
In contrast, in our dataset, the norms are more general and binned to a finite set, which is more tractable to train Value-Aligned agents.
Some other works on representing social norms and values over everyday situations in natural language are SCRUPLES~\cite{lourie2020Scruples}, ETHICS~\cite{hendrycks2020ethics} and SOCIAL CHEMISTRY 101~\cite{forbes2020social}.
SCRUPLES collected 32k real-life anecdotes with normative judgments from a subreddit forum to construct the dataset, where \cite{forbes2020social} is a larger corpus on social norms and moral judgment consisting of 292k annotated situations.

%Delphi~\cite{jiang2021delphi} has combined the corpora from these previous norms and values datasets (ETHICS~\cite{hendrycks2020ethics}, Moral Stories~\cite{emelin-etal-2021-moral}, SOCIAL CHEMISTRY~\cite{forbes2020social} and  SOCIAL BIAS INFERENCE CORPUS~\cite{sap2020social}) and created an unified large dataset on social norms and ethics named as COMMONSENSE NORM BANK.
Delphi~\cite{jiang2021delphi} has combined the corpora from ETHICS~\cite{hendrycks2020ethics}, Moral Stories~\cite{emelin-etal-2021-moral}, SOCIAL CHEMISTRY~\cite{forbes2020social} and  SOCIAL BIAS INFERENCE CORPUS~\cite{sap2020social} and created a unified dataset on social norms and ethics called the COMMONSENSE NORM BANK.
Trained on this unified dataset, Delphi is able to make moral judgments in real-life situations.
However, it is unable to provide the notion of social norms or principles based on which the judgment was made. Moreover, the compiled corpus used in Delphi was originally collected from online forums such as Reddit.
% and annotated by crowd workers who are not necessarily experts on moral judgment. 
%forum, i.e., subreddit AITA, and annotated by crowdsource workers who are not experts on moral judgment.
Thus, it may contain an inappropriate and biased set of examples, which can lead to improper and biased moral decisions by the model trained with this data.
In contrast, we present the corpus collected from children's stories, which are deliberately meant to teach social norms to children. We further curated the corpus by including only recent stories, ensuring the dataset quality and its alignment with recent societal norms. Moreover, the dataset is multimodal, with an image associated with each natural language story example.

\section{Dataset}
To facilitate Value Alignment, we built a multimodal dataset, the \GG{} story corpus, illustrating social values through natural language texts and images. 
The dataset contains illustrations of social behaviors labeled as normative or non-normative and provides the inherent social principles or values of these behaviors as well.
Thus, the dataset consists of two sub-datasets: the GnG Normative dataset, which describes normativity, and the GnG Principles Dataset, which describes the underlying social principles of involved in each example. 
% Thus, the dataset consists of two sub-datasets: one describes normativity, and the other describes the underlying social principles of involved in each example. 
We have constructed this dataset using a Children's comic strip named \GG{}. In this section, we discuss the methodology employed to create these two dataset components in detail.

\subsection{GnG Normative Dataset}
%To facilitate the task of Value Alignment, in this work we built a multi-modal dataset illustrating social values through natural language texts and images.  normative behavior using a children comic strip \GG{}. The \GG{} comic strip is published by Highlights Magazine that begin its run in 1946.
The dataset, built for the aid of the value alignment task, is based on a Children's comic strip named \GG{}.
The \GG{} comic strip is published by Highlights Magazine and began its run in 1946.
It is meant to convey societal values to young children by providing them with examples of desirable and undesirable behaviors. 
These behaviors were depicted using the two titular main characters of the strip: Goofus and Gallant. 
Goofus and Gallant are young boys who have specific character traits. 
Goofus is a child that typically performs undesirable actions. 
The implication is that the behaviors that Goofus performs are not meant to be emulated. 
In contrast, Gallant is a young boy who typically performs desirable, socially acceptable actions that are meant to be emulated. 
This setup provided us with an automatically labeled corpus where all actions done
by Gallant are labeled as normative, and all actions done by Goofus are labeled as non-normative. We named this dataset the GnG Normative dataset.
% Each comic is composed of two comic panels with text associated with each panel. 

The advantage of \GG{} comic is that it has both image and text for each strip demonstrating an action in a social scenario. 
% Thus, we can utilize both image and text information to identify societal norms. 
This allows us to utilize both visual and textual information to identify societal values.
To better ensure that the machine learning models can learn relevant social values, we collected the most recent strips from 1995-2017. 
We extracted the text from each strip's panel, but as the older images are of a lower visual quality than the newer images, we only included images from 2001 to 2017. 
This provided us with 1387 texts and 819 images. 

As a result, we created two versions of this dataset: GnG text-only and GnG multi-modal normative dataset.
The text-only version contains 1,387 texts collected from the comic panels. 
The multi-modal version includes the 819 images along with their corresponding texts from the comic panels. 
Additionally, we removed any explicit references to Goofus and Gallant by replacing their names with pronouns such as "he" or "they".
% We also removed any explicit references to Goofus and Gallant by replacing their names with pronouns such as “he” or “they”.

\subsection{GnG Principles Dataset} 
%The GnG normative dataset we have created is the corpus to categorize an action into normative or non-normative. 
The GnG normative dataset we created serves as the corpus for categorizing actions as normative or non-normative.
It does not have any identifying information that expands on what principle or value is contained in each comic. 
But along with labeling actions as socially acceptable/unacceptable, it is crucial to know which social norms or principles are violated or upheld by these actions. 
To address this, we extended the GnG normative dataset by annotating Goofus and Gallant's actions with relevant social values or principles, creating the GnG Principle dataset. 

 % A comprehensive description of the data collection process is provided in the following section.
% \subsubsection{Data Collection}
%The objective of the data collection task is to use human annotators to annotate the \GG{} comic strip with social principle information to generate a corpus of normative social principles. 
% The objective of the data collection task is to annotate the \GG{} comic strip with social principle information, creating a corpus of normative social principles.
Before beginning the annotation task, it is essential to determine how annotators will provide the social principle information. 
% To systematically annotate the \GG{} comic strip with social principle information, it is essential to determine how annotators will provide this information. 
We considered two approaches: 1) Writing the principles in a free-form text and 2) Selecting from a given set of social principles. 
The free-form texts will make the unique set of principles very diverse, which is difficult to generalize. 
Therefore, to make the principle classification problem more tractable, we decided to restrict the choice of principles to a finite set. 
For this purpose, we utilized the system introduced in~\cite{kiesel-etal-2022-identifying} to define the set of ``social principles". 
In their work, Kiesel et al. proposed a value taxonomy with 54 values, which are both relevant and supported by social science research. 
We further downsized the number of values to better align it with the action description of \GG{} corpus. 
We ran the pre-trained value model developed by Kiesel et al. on the \GG{} dataset to get the zero-shot value prediction on the text descriptions of the actions in the corpus. 
This experiment gave us 27 social values aligned with the \GG{} texts.
We considered these 27 social values to be the set of predefined principles that would label each action of Goofus and Gallant.

For this annotation task, we utilized both the text descriptions of the actions and their corresponding images. Therefore, we only annotated the instances from the GnG multi-modal dataset where images are available for the corresponding textual descriptions of the actions. We ran two data collection processes to curate this dataset: 1) using crowdsource workers and 2) utilizing Large Language Models (LLMs). In the following, we describe each process in detail.

\subsubsection{Data Curation--Using Crowdsource Workers}
% asked the crowdsource workers to annotate the given image-text pairs by selecting the best representative values from this list that are upheld or violated by the actions described in the image-text pair. 

In our first attempt to build the GnG Principles dataset, we used crowdsource workers to annotate the principles. We recruited annotators exclusively from English-speaking countries.
In the task, the annotators were given image-text pairs from the GnG multi-modal dataset and a fixed set of social principles curated through the process we discussed earlier.
The images provided a visual illustration of the scene, while the associated text described the action performed in the scene.
For each given data item, the annotators were required to provide the three most representative principles from the given principles list that were upheld or violated by the action demonstrated in the corresponding image-text pair. 
Each data item was annotated by three different annotators, with each annotator labeling eight items from the corpus.
We recruited 150 crowd workers and annotated 400 examples, which is approximately half of the total number of examples.

After collecting responses from the annotators, we computed the frequency of each principle selected by the annotators for every data instance. The principle that appeared most frequently for a given instance was chosen as the final label, ensuring that the most commonly agreed-upon principle was used.
To assess the quality of the collected annotation, we computed the inter-annotator agreement using the Fleiss-Kappa score~\cite{fleiss1971measuring}. The resulting kappa score was 0.54, indicating a moderate level of agreement among the annotators. To understand the reason behind this moderate agreement, we investigated further and found that very often, different annotators selected varying principles for the same example. This variability arose because social principles can have different meanings to different individuals and the same action can be aligned with multiple principles as well.

This variability made the class labels of the dataset sparse, which is usually difficult for machine learning models to learn from. To address this challenge, we sought an alternative approach to curate the dataset. We explored utilizing the capabilities of Generative AI, which is our second attempt to annotate the data, aiming to provide a more consistent and robust set of annotations. 
% During crowd-sourced annotation, we observed that different annotators selected varying principles for the same example. This variability arises because social principles can have different meaning to different persons and the same action can satisfy multiple principles as well.
% So, the collected set of principles still remains sparse and hard to generalize.
% However, we checked the quality of our collected annotation by evaluating the inter-annotator agreement with the Fleiss kappa~\cite{fleiss1971measuring} score. 
%As data is multi-label and there were more than two annotators for each data item, we consider the Fleiss kappa~\cite{fleiss1971measuring} score as the metric of the inter-annotator agreement and the score is 0.49. 
%We removed the annotations where the annotators could not reach an agreement on any label. After removing such annotations, the Fleiss kappa score of the annotations becomes 0.54. 
%\subsubsection{Data Collection using GPT-4o}
% As the Fleiss kappa score for the crowd sourced principles annotation came out moderate $(0.54)$ based on~\cite{landis1977measurement}, this indicates that the task is inherently challenging and selected principles are infrequent. 

\begin{figure}
\centering{
\includegraphics[width=0.8\columnwidth]{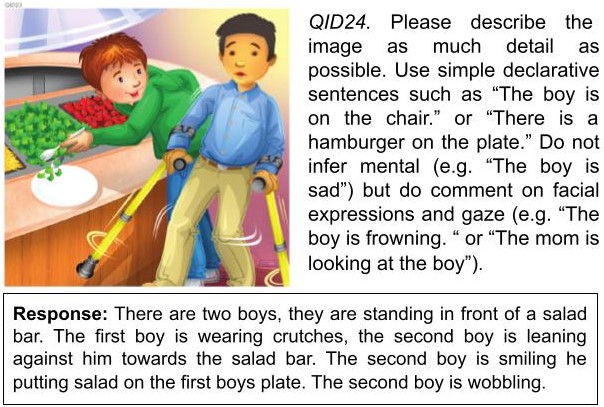}} % Reduce the figure size so that it is slightly narrower than the column.
\caption{Collecting scene description: An example image provided instructions and response collected from a crowdsource worker.}
\label{fig:image_description}
\end{figure}

\subsubsection{Data Curation--Utilizing Large Language Models}
% So, we alternatively investigated a different way to curate the data collection process. 
% Our investigation incorporated Generative AI's capabilities for the annotation task. 
% We used the SOTA GPT-4o~\cite{openai2024gpt4} in this process. 
To create a more comprehensive principles dataset, in this process we utilized the capabilities of LLMs to annotate the Goofus and Gallant actions with human values. 
% Specifically, we used the state-of-the-art GPT-4o model~\cite{openai2024gpt4} from OpenAI. 
Specifically, we used the API provided by OpenAI to access the pre-trained LLMs. 
Since LLMs support text-only prompts, we included the textual description of the images instead of the actual images. To collect these textual descriptions, we involved crowdsource workers. Therefore, the entire principle annotation process consisted of three steps: first, collecting detailed descriptions of the scene illustrated in the images; second, using LLMs to predict principle labels for each data point, and finally, verifying the correctness of the responses of LLMs by humans. The details of each step are described below.
\paragraph{Collecting Scene Descriptions}
In this step, we collected detailed descriptions for the images in the GnG multi-modal dataset. To ensure high-quality descriptions, we employed human annotators to write them. We provided the annotators with detailed instructions outlining the criteria for writing the descriptions.

The instructions specified that descriptions should be written in simple declarative sentences and needed to include the state of the different characters in the image and their interactions, as well as their facial expressions or gaze. Additionally, the descriptions were required to detail the objects in the environment and their properties, as well as the actions taking place. This structured approach ensured that the descriptions were comprehensive and consistent, providing a rich textual representation of the images for further annotation. Figure \ref{fig:image_description} shows an example of the collected response for an image.

For each image, we employed two annotators. The first annotator created the initial description based on the provided criteria. Afterward, the second annotator reviewed the description, adding any missing information if necessary to ensure all relevant details were captured. This two-step process improved the quality and comprehensiveness of the scene descriptions.

% The instructions specified that descriptions should be written in simple declarative sentences, such as “the boy is on the chair.” Each description needed to include the state of the different characters in the image and their interactions, as well as their facial expressions or gaze, for example, ``the person is frowning'' or``the dad is looking at the boy.'' However, annotators were instructed not to infer mental states, such as ``sad'' or ``excited.'' Additionally, the descriptions were required to detail the objects in the environment and their properties, as well as the actions taking place, such as ``the boy is walking.'' This structured approach ensured that the descriptions were comprehensive and consistent, providing a rich textual representation of the images for further annotation.

% We provided 3 inputs in the prompt of the GPT: the scene description, action description and compliance inoformation. 
% The scene description had been collected on \GG{} comic images using crowd source workers. 
% The description contains simple declarative sentences like ``The boy is on the chair" and facial expressions or gaze like ``The person is frowning" or ``The dad is looking at the boy", but no mental state like ``sad" or excited". 
% The action description is directly collected from the text of each comic strip. 
\paragraph{Annotating Principles using LLMs}
After collecting the scene descriptions, we applied the zero-shot prompting technique of Large Language Models (LLMs) to predict the principles upheld or violated by the actions of Goofus and Gallant. We used the state-of-the-art GPT-4o model~\cite{openai2024gpt4} from OpenAI.
% After collecting the image descriptions, we used the zero-shot prompting technique of LLMs to predict the principles of the GnG actions that have been violated or upheld by the actions. We used the state-of-the-art GPT-4o model~\cite{openai2024gpt4} from OpenAI for our predictions. 
We provided three inputs in the prompt to the GPT model: the scene description, action description and compliance information. 
The compliance information indicates whether the action follows or violates social norms or principles, labeled as either \textit{followed} or \textit{violated}.
If the comic strip text features Gallant, the compliance information is set as \textit{followed}, and for Goofus, it is \textit{violated}.
% In this work, we are annotating the social principles rather than labeling the actions as normative or non-normative.
% Therefore, we have shared the character names to enable GPT to learn from the inputs and effectively determine the principles.

We applied the chat format for the prompt, where we provided an elaborated description of the task, our predefined set of principles along with their definitions, instructions to the GPT model, and the expected output format in the system prompt. Then, in the user prompt, we provided our inputs for the GPT model.
We instructed the GPT model to identify at least two principles from the predefined list mentioned in the prompt for the given action.
The first principle generated should be the most representative one, and the second generated principle should be the second most. We refer to the first principle as Principle 1 and the second principle as Principle 2.
If the action does not align with any of the principles listed, the model can suggest a new principle. 
Furthermore, the GPT model has been asked to provide detailed explanations analyzing how the selected principles represent the given action. 
These explanations are generated for each principle assgined by the LLM. 

For each example, we queried the GPT model five times. This approach enabled us to determine which principles were predicted most frequently across these queries. When the same principle was predicted multiple times for a given input, it indicates higher confidence from the model in that prediction.
Thus, to ensure accurate and reliable annotations, we selected the most frequently predicted principle from these five iterations as the final prediction for Principle 1.
If two principles had the same frequencies, then we randomly selected any one of them as the most frequent one.
For the Principle 2, we also selected the most frequent one. But if the most frequent one had already been selected as Principle 1, we chose the second most frequent principle for the second category.

\paragraph{Human Verification}
After post-processing the GPT responses and finalizing the predictions for principle 1 and principle 2, we evaluated the correctness of these predictions through human review. For each instance, we provided the scene description, action description, compliance information, and the two principles predicted by GPT to human reviewers. The reviewers were asked to select the principle they believed was violated or upheld by the action. If both principles seemed applicable, they were instructed to select both. If neither of the predicted principles was correct, they were to select ''None.''

We initially ran this process on 50 instances of the GnG multi-modal dataset to evaluate how well the responses of LLMs aligned with human judgment. We employed two annotators for each instance, with each annotator reviewing 10 instances, totaling 10 annotators for the initial review. After collecting their responses, we computed their agreement with the LLM predictions and found the agreement rate is 91\% between 2 annotators for atleast one principle. This high level of agreement indicated that the GPT-4o predictions were of high quality and aligned with human judgment. Consequently, we extended the data collection process to the remaining instances of the GnG multi-modal dataset. For quality assurance, we randomly selected 100 instances from the extended dataset and reviewed them using the same protocol. We achieved a 93\% agreement rate where the annotators agreed with at least one principle and a 61\% agreement rate where the annotators agreed with both principles predicted by GPT-4o. 

% By selecting the most frequently predicted principle, we aimed to ensure a more accurate and reliable final annotation.

% Collecting two most relative social values and explanations for them on 800 text examples of \GG{} dataset using GPT-4o is done five times. 
% After five iterations of data collection, we gathered the generated principles under each output category and measured the number of frequencies for each principle.
% For finalizing the first principle category which holds the most relevant social value, we selected the principle which was predicted most.
% In case of principles generated for the $2^{nd}$ output principle, if the principle is already determined for the 1st output principle, we chose the $2^{nd}$ frequent principle or randomly selected one of the principles if they have same frequency. 

% As in this work, we are annotating the social principles and not labeling the normative or non-normative actions, we have shared the character names to let the GPT learn from the inputs and decide the principles effectively. 

A summary of each dataset used in our experiments can be found in Table~\ref{tab:tabledata}.

\begin{table}[tb]
\centering
\footnotesize
\caption{Dataset summaries.} %\textit{Hand-selected} refers to the count after manual review. \textit{Consensus} refers to the count after filtering events with significant MTurk annotator dissent. }
\renewcommand{\arraystretch}{1.4}
\label{tab:tabledata}

\begin{tabular}{|l|c|c|c|c|}
% \multicolumn{1}{c}{} &
% \multicolumn{1}{c}{} &
% \multicolumn{1}{c}{} &
% \multicolumn{1}{c}{} \\
\hline
\multicolumn{1}{|c}{\textbf{Dataset}} &
\multicolumn{1}{|c}{\textbf{Modality}} &
\multicolumn{1}{|c}{\textbf{Total}} &
\multicolumn{1}{|c}{\textbf{Train}} &
\multicolumn{1}{|c|}{\textbf{Test}} \\
\hline
\multirow{2}{*}{GnG Normative} & Text only &$1387$&832&555\\
 & Multi-modal &$819$&$655$&$164$\\
\hline
\multirow{2}{*}{GnG Principles} & Text only  &$819$&$655$&$164$\\
&(with scene descriptions)&&&\\
\hline
\end{tabular}
\end{table}

% \begin{figure}
% \centering{
% \includegraphics[width=1.0\columnwidth]{figures/nvnn.png}} % Reduce the figure size so that it is slightly narrower than the column.
% \caption{Examples of test dataset text. }
% \label{fig:nvnn}
% \end{figure}

\section{Tasks}

We aimed to show that the stories in the \GG{} dataset contain rich knowledge about socio-cultural norms and values that can be learned by machine learning models. 
To achieve this, we explored two classification tasks on the \GG{} dataset. In the first task, we investigated whether we could determine if the action described in the story is socially acceptable or not (normative or non-normative). In the second task, we examined whether we could identify the social principle that is followed or violated by the action described in the story. Formally the two tasks are:
\begin{itemize}
    \item Normativity classification
    \item Principles classification
\end{itemize}

% These two tasks will allow us to verify how well the stories of the G\&G dataset contain the knowledge of social values and norms. We have implemented several classification models to conduct each task. 
% As GnG dataset contains text data that describes an action taken by the subject of the story, we utilized this text information for the classification tasks. 

%G\&G dataset contains multi-modal data where each data item consists of an image and an associated text that describes an action taken by the subject of the story. The multi-modal property of the dataset allows us to utilize both image and text information for the classification tasks.
\subsection{Normativity Classification}
In this task, the objective is to classify the actions demonstrated in the \GG{} comic strips as either normative or non-normative. 
We seek to show that knowledge of socially normative and non-normative behavior can be identified from naturally occurring stories. 
To achieve this, we conducted experiments to develop baseline machine learning models that can classify normative behavior from both textual descriptions and visual illustrations.

\subsection{Principles Classification}

% Along with categorizing behaviors into either normative or non-normative, it is also important to know the inherent social principles of these behaviors.
% This knowledge will help agents and humans learn the violated or upheld social principle behind the action, aiding in identifying and correcting misclassifications of normative behavior.
% This knowledge will allow both agents and humans to learn the underlying social principle or value which has been obeyed or violated by the action and help to better understand the reason or even remedy the misclassification of normative behavior. 
In this task, we aimed to develop systems capable of understanding descriptions of human behavior with respect to normative principles. We defined normative principles as the guidelines that direct individuals to adhere to a society's collective behavioral rules, such as ``be respectful to traditions'' or ``be responsible''.
%We defined normative principles as the set of principles that guide people to conform to a collective set of behavioral rules that a society adheres to. Examples of potential normative principles might be “be behaving properly” or “have an objective view”. 

For this classification task, we used the GnG Principles dataset, created by augmenting the GnG multi-modal dataset with principles collected using the GPT-4o model and subsequently verified by humans. 
We trained several machine learning models to predict the inherent social principles underlying the actions described in the text. 
Usually, social behaviors or actions can comply with or violate multiple social values concurrently.
For instance, “Gallant does his studying before watching TV”, can comply both normative social principles like “be responsible” and “be compliant”. 
Because of this inclusive property of social norms, we evaluated the correctness of the classifiers differently from the standard multi-class classification. In this task, a predicted principle is considered correct if it matches either of the two true principles.
% Because of this inclusive property of social norms, we framed this task as a multi-label multi-class classification problem. 
% For each text description, the objective of the classifier is to predict the top three representative principles that are being upheld or violated by the behavior described in the text.

\begin{figure}

  \centering
  \includegraphics[width=0.99\columnwidth]{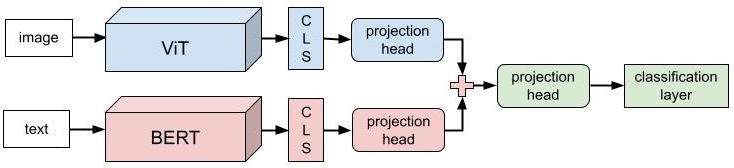} % Reduce the figure size so that it is slightly narrower than the column.
  \caption{Network architecture of the image-text dual encoder GnG model }
  \label{fig:dual_encoder}
\end{figure}%

\begin{table*}[tb]
\centering
\footnotesize
\caption{Results of the Baseline models trained and tested on GnG Multi-modal Normative dataset} 
\renewcommand{\arraystretch}{1.4}
% \begin{tabularx}{\columnwidth}{|p{1.2cm}|p{1.4cm}|X|X|X|X|X|}
\begin{tabular}{|l|c|c|c|c|c|c|}
\hline
\textbf{Modalities} &
\textbf{Model} &
\textbf{Acc} &
\textbf{F1} &
\textbf{Precision} &
\textbf{Recall} &
\textbf{MCC} \\
\hline
Image & ViT-CL &70.32&72.94& 78.48 &68.13 & 40.94\\
& ViTForImageClassification & 72.26 & 77.25 & 74.49 & \textbf{80.22} & 42.02\\
\hline
Text &
BERT-CL & 70.97 & 74.86 & 76.14 & 73.63 & 40.56\\
&
BertForSequenceClassification & 72.90& 77.17& 76.34& 78.02& 43.87\\
\hline
Image \& Text & Dual Encoder & \textbf{76.77} & \textbf{79.78} & \textbf{81.61} & 78.02 & \textbf{52.61}\\
\hline
\end{tabular}
\label{tab:results}
\end{table*}

\section{Baselines}

We built several classifiers for each of the tasks to determine the best-performing models for these tasks. 
We made use of transformer-based vision and language models to implement the classifiers.
% We made use of transformer-based models to implement the classifiers.
In this section, we discuss the details of these models.
\subsection{Normativity Classification Models}
Using the images and texts of the GnG Normative dataset, we trained multiple binary classifiers capable of classifying events in stories as normative or non-normative. 
First, we built a model using only images as input to investigate the effectiveness of visual context in determining normativity. Next, we used only texts as input for the model. Finally, we incorporated both text and images as inputs in the model. In this way, we comprehended the implications of different modalities in classifying normativity.
% To classify normativity, we built a binary classifier based on the text input data.
% This classification task helps us to understand the effect of text snippets on classifying normativity.

% To classify normativity, we built three binary classifiers based on the modality of the input data.
% At first, we built the normative model using only the images as input to investigate how well the visual context is useful to determine values information. 
% In the second model, we injected text along with the image to examine the influence of text in values identification. 
% And as the third model, we only used the text snippet as the input.
% It will help us to understand the effects of different modalities on classifying normativity.

\subsubsection{Image Only Model}
In this model, we used only images as input to classify actions into normative or non-normative classes. We implemented two binary classifiers using the pre-trained Vision Transformer (ViT)~\cite{ViT2020}: 1) ViT with Custom Layers (ViT-CL) and 2) Pre-trained Image Classifier (ViTForImageClassification).
For the first model, we used ViT as the base model and added a projection layer followed by a classification layer on top. The projection layer consists of a linear layer, an activation function, dropout, and layer normalization. The ViT base model provides the embedding of the special [CLS] token, which represents the entire input image. We passed this embedding vector through the projection and classification layers to make the final prediction.

In addition to the pre-trained base ViT, there are off-the-shelf ViT models specifically trained for image classification tasks. For our second classifier, we fine-tuned one of these models, ViTForImageClassification, without adding any additional layers since it already includes a trained classification layer.

\subsubsection{Text-based Models}
The text-based classifiers take sentences as input and predict whether the event described is normative or non-normative. Similar to the image-only model, we implemented two different binary classifiers for the text-only input: 1) BERT~\cite{devlin-etal-2019-bert} with Custom Layers (BERT-CL) and 2) Pre-trained Sequence Classifier (BertForSequenceClassification).

The first classifier we implemented is a transformer-based large language model with a similar projection head and classification layer added on top. The language model provides the contextualized embedding of the input sentence, represented by the embedding vector of a special classification token [CLS]. This embedding vector is then passed through the projection and classification layers to make the prediction. 

For the second text-only model, we used a transformer-based pre-trained sequence classification model, BertForSequenceClassification, without adding any additional layers to leverage its capabilities directly.

% Along with the pretrained base large language models, there are off-the-shelf language models that have been pretrained for different downstream tasks such as question answering and sequence classification. For our second text-based classifier, we fine-tuned such off-the-shelf pretrained sequence classification models without adding any additional layers to leverage their capabilities directly.

\subsubsection{Image and Text model}
To investigate how visual and textual information concurrently influence the classification of normative and non-normative actions, we trained a binary classifier that uses both image and text as inputs and implemented a transformer-based dual encoder network for it.

We combined the previously implemented ViT-CL and BERT-CL models using a projection head and a classification layer to build this model. The ViT-CL extracts the embedding vectors for images, while BERT-CL provides the embedding vectors for texts. The projection head aligns both image and text embeddings into the same latent space. Finally, the classification layer, which includes a linear layer followed by a softmax function, makes the prediction.
Figure~\ref{fig:dual_encoder} shows the model’s architecture. 

\subsection{Principles classification Models}
To build the classifier for principle prediction, we utilized text-based information as the input. We fine-tuned a pre-trained sequence classification model, similar to the one used in the Text-only Normativity Classification Models mentioned previously, to implement the models.
% To implement the models, we fine-tuned a similar pre-trained sequence classification model employed in the Text-only Normativity Classification Models mentioned previously.

For this task, we used the GnG Principles dataset to train the models. We provided the models with three inputs: the scene description, the action description, and the compliance information. These inputs allow the model to understand the environment of the scene, the specific action taking place, and whether the action complies with or violates social values. The model then predicts the principle that is either followed or violated by the action described in the scene. By leveraging the rich contextual embeddings generated by the transformer-based language models, our approach aims to accurately identify the underlying social principles guiding the actions within the stories. 

%  To classify principles, we have used text descriptions as input information.
%  As mentioned earlier, we frame the principles classification problem as a multi-label multi-class classification problem. For each text description, the objective of the classifier is to predict the 3 most representative principles. To implement the classifier, we utilize the transformer based pre-trained large language model. On top of the transformer model, we added a classification layer that consists of two fully connected (FC) layers. We took the vector representation of the [CLS] token of the LLM which represents the embedding vector of the input text. This embedding vector is then passed to two fully connected layers to make the final prediction on the text description. As the classifier needs to predict multiple classes for each text, we applied sigmoid activation function on the output of the final FC layer. 
% Though sigmoid activation is usually used for binary classifications, we used this activation function in our multi-class principle classifications problem as there are multiple correct labels for each input. The sigmoid function provides probability for each class separately. Therefore, we can compare these class probabilities with true class label (one hot encoding) to calculate the loss of the network to train it. 
% During inference, to determine the predicted classes, we defined a threshold probability value. All the classes with a probability higher than the threshold are considered as the predicted classes and thus the rest of the classes are skipped. 

\begin{table}[tb]
\centering
\footnotesize
\caption{Comparing the full Text-based models and sentiment analysis model on the test data of GnG text-only Normative Dataset.} 
\label{tab:sent_model}
\renewcommand{\arraystretch}{1.4}
\resizebox{\columnwidth}{!}{%
\begin{tabular}{|l|c|c|c|c|c|}
\hline
\multicolumn{1}{|c}{\textbf{Model}} &
\multicolumn{1}{|c}{\textbf{Acc}} &
\multicolumn{1}{|c}{\textbf{F1-score}} &
\multicolumn{1}{|c}{\textbf{Precision}} &
\multicolumn{1}{|c}{\textbf{Recall}} &
\multicolumn{1}{|c|}{\textbf{MCC}} \\
\hline
BERT & \textbf{94.05}& \textbf{94.03}& \textbf{94.55}& \textbf{93.53}&\textbf{88.11}\\
DistilBERT & 93.69 & 93.65 & 94.51& 92.81 & 87.40 \\
Sentiment Analysis & 62.46 & 58.44 & 66.81 & 51.94 & 25.83 \\
\hline
\end{tabular}
}
\end{table}

\begin{table}[tb]
\centering
\footnotesize
\caption{Ablation Studies performed for Principles Classification task. Bold and underlined numbers show the best scores for the relaxed and strict metrics, respectively.} 
\label{tab:principle}
\renewcommand{\arraystretch}{1.4}
\resizebox{\columnwidth}{!}{%
\begin{tabular}{|p{1.2cm}|p{1.2cm}|c|c|c|c|c|}
% \begin{tabular}{|p{1.5cm}|c|c|c|c|c|}
\hline
\textbf{Input} &
\textbf{Evaluation Type} &
\textbf{Acc} &
\textbf{F1-score} &
\textbf{Precision} &
\textbf{Recall} &
\textbf{MCC} \\
\hline
Scene +  & Strict & 52.26& 16.84& 18.06&18.13&35.94\\
 Action & Relax & \textbf{74.84} & 32.76 & 32.64& 34.04 & \textbf{63.72}\\
\hline
\multirow{2}{*}{Action}  & Strict & \underline{52.90} & \underline{19.15} & \underline{20.21} & \underline{19.57} & \underline{37.00}\\
& Relax & \textbf{74.84} & \textbf{36.06} & \textbf{35.18} & \textbf{37.20} &63.53\\
\hline
\end{tabular}
}
\end{table}

\section{Experiments and Results}
To show the effectiveness of our GnG Normative and GnG Principles dataset, we conducted two sets of experiments: 1) Normativity Classification experiments and 2) Principles Classification experiments. We used accuracy, precision, recall, f1-score and matthews correlation coefficient (MCC) as the evaluation metrics to assess classification quality.

% New result
% text only: validation Loss: 0.0377, Validation Acc: 68.7500, precision: 0.7614, recall: 0.7363, f1-score: 0.7486, accuracy: 0.7097, mcc: 0.4056
% text and image result on 771 data: validation Loss: 0.0349, Validation Acc: 74.3750, precision: 0.8161, recall: 0.7802, f1-score: 0.7978, accuracy: 0.7677, mcc: 0.5261

% \begin{table*}[tb]
% \centering
% \footnotesize
% \caption{Results} 
% \label{tab:results}
% \begin{tabular}{|l|c|c|c|c|c|c|}
% \hline
% \multicolumn{1}{|c}{\textbf{Modalities}} &
% \multicolumn{1}{|c}{\textbf{Model}} &
% \multicolumn{1}{|c}{\textbf{Test accuracy}} &
% \multicolumn{1}{|c}{\textbf{f1 score}} &
% \multicolumn{1}{|c}{\textbf{Precision}} &
% \multicolumn{1}{|c}{\textbf{Recall}} &
% \multicolumn{1}{|c|}{\textbf{MCC}} \\
% \hline
% Image &
% Zero-shot
% &$0.535$ &$0.446$ & $0.744$ & $0.319$ &\\
% & Fine-tuned
% &$0.677$&$0.725$&$0.725$&$0.725$ &\\
% \hline
% Image \& Text &
% Dual Encoder &
% &$1462$&$900$&$555$ & \\
% & VisualBERT & & & & &\\
% \hline
% Text &
% DistilBERT & 0.716 & 0.778 & 0.72 & 0.846 & 0.402\\
% &
% DPCNN & & & & &\\
% &
% BERT & & & & &\\
% &
% XLNet & & & & &\\
% &
% T5 & & & & &\\
% \hline
% \end{tabular}
% \end{table*}

\subsection{Experiment 1: Normativity Classification}
In this experiment, we seek to identify how well the machine learning models can classify normative and non-normative behavior from unseen Goofus and Gallant texts and images when trained on the GnG training set. It will allow us to understand how well the machine learning models can extract value information from the GnG normative dataset. 

We conducted ablation studies to investigate the input modalities that contribute the most to classifying normative behavior and the best-performing machine learning models on these modalities. The results of this experiment are shown in the Table~\ref{tab:results}. It shows the performance of different models on classifying normative behavior from different input modalities. All models in this experiment were trained and tested on the GnG multi-modal normative dataset. Additionally, we used the full GnG text-only normative dataset to train text-only models. 
% Despite this, the fine-tuned Vision Transformer(ViT) model performs significantly better than the zero-shot ViT. 
% We utilized a pre-trained Vision Transformer (ViT) to classify the normative/non-normative behavior from images to investigate the performance of the zero-shot transfer without further training. We have seen, though the zero-shot transfer could not detect the classes very accurately, fine-tuning the model with \GnG{} training images improves its performance substantially. It indicates that the images of \GnG{} comic strip possess meaningful social values information and can be used to identify normative social behavior. 

From the table~\ref{tab:results}, both the image and text-only models achieve test accuracies greater than 70\%, with the MCC exceeding 40. This indicates that both \GG{} images and texts contain value information that can be extracted by state-of-the-art machine learning models. Combining images and text in the input significantly improves the performance of predictions. The Dual Encoder, which combines ViT-CL and BERT-CL to process both image and text inputs, outperforms all other models that use a single modality.

The table shows that the fine-tuned ViTForImageClassification model slightly outperforms the custom ViT-CL model. A similar trend is observed in the text-only models as well, where BertForSequenceClassification performs better than the BERT-CL. One potential reason could be that the classification layers of the ViTForImageClassification and BertForSequenceClassification models have already been pre-trained with ample data, whereas the additional custom layers in ViT-CL and BERT-CL were only trained on data from the GnG dataset.

The models discussed above were trained exclusively using the images and texts from the GnG multi-modal normative dataset. Given the availability of additional text data in our GnG text-only normative dataset, we trained another set of text-only models on this complete dataset to further explore the effectiveness of our text-based data in classifying normativity from action descriptions. Table~\ref{tab:sent_model} displays the performance of these models. Both the BERT and DistillBERT models achieve accuracies and f1-score exceeding 93\%, which further validates the quality of the \GG{} dataset

Finally, we ran a pre-trained sentiment analysis model on the GnG text-only test data to inspect if the sentiment analysis and classification of normative behaviors tasks share similar observations. We use the transformer-based pre-trained model \textit{distilbert-base-uncased-finetuned-sst-2-english}~\cite{sentiment_url} from Hugging Face for the sentiment analysis, which was trained on Stanford Sentiment Treebank~\cite{sentiment2013socher} corpora and is one of the state-of-the-art models in this task. The experiment shows that the pre-trained sentiment analysis model struggles to detect normative/non-normative behaviors from the texts. It indicates the two tasks are not analogous to one another and require two different types of data to work with. 
\subsection{Experiment 2: Principles Classification}
Principles classification is a multi-class classification task that aims to label each example of GnG Principles dataset with the two most representative principles that best identify social values illustrated in the action text.
Since the dataset is textual, we used the BertForSequenceClassification model, which we fine-tuned and tested on the dataset.
We performed 2 ablation studies based on the inputs we used. 
One study employed both scene and action descriptions, while the other used only action descriptions. 
Table~\ref{tab:principle} shares the results of the experiment done for the classification task. 
As the compliance information was the common input for both studies, we did not add it to the table. 

Recall that each example in the GnG principles dataset has two target principles: Principle 1 (the most representative) and Principle 2 (the second most representative). Since each example has two correct answers, we evaluated the model's performance using a slightly different approach than the traditional method. We decided the correct prediction in two separate ways: 1. Strict - the predicted principle is correct if it matches Principle 1 (the most representative principle). 2. Relax - the predicted principle is correct if it matches either of the two true principles.

Table~\ref{tab:principle} shows the results of this experiment. We trained two models based on the input information. In the first model, we included both scene and action descriptions along with compliance information. In the second model, we provided only the action description with compliance information. This allowed us to investigate the influence of scene descriptions on predicting the principle of the action and to determine if the model could identify the principle information from the action description alone without the scene description. 
From the results, we can see that the classification model using only action descriptions performed better in both the strict and relaxed metrics.

% We observed that the classification model using only the action description as input performed better than the model using both action and scene descriptions as input. This is consistent for both strict and relaxed evaluation measures. We hypothesize that the scene descriptions are much longer than the action descriptions, providing a lot of additional information that may not always be directly relevant to identifying the principles. This extraneous information can overwhelm the model, making it harder to focus on the important information necessary for accurate classification. While more sophisticated models, such as instruction-tuned LLMs, can be guided to prioritize relevant information, our simpler baseline models lack this capability. This suggests there is significant scope to improve the performance of these models in the future

\section{Discussion}
In this study, we present a multi-modal dataset to facilitate value alignment. Our comprehensive experiments demonstrate the effectiveness of this dataset. From the first experiment, we observe that the normativity classification models can effectively distinguish between normative and non-normative actions of the GnG dataset.
%, particularly the text-only model with the full-text dataset, which achieves an F1 score of 94.03. 
However, some instances are difficult to classify. For example, “I think I broke your camera, Dad. I'm sorry, says he”. In this scenario, the boy's action of breaking the camera is not normative, but his apology instead of hiding the matter is commendable. In such instances, our model struggles, as these are particularly difficult examples. 
%This highlights the challenge of the normativity classification task. We believe that further improvements over our baseline models are possible with the use of more advanced models.}

In our experiments of principles classification, we observe that the classification model using only the action description as input performed better than the model using both action and scene descriptions as input. This is consistent for both strict and relaxed evaluation measures. We hypothesize that the scene descriptions are much longer than the action descriptions, providing a lot of additional information that may not always be directly relevant to identifying the principles. This extraneous information can overwhelm the model, making it harder to focus on the important information necessary for accurate classification. 

\section{Conclusion}

%Normative actions are the behaviors that a society expects to be followed for corresponding situations. It is beyond good and bad behavior. Normativity depends on the situation and is attached to the norms and values of a society; thus, the task of classifying normative behaviors differs from classifying positive and negative actions. 
While value alignment is a noble goal, many subtleties surrounding values must be addressed before any practical alignment can be achieved. 
While there has been much research on utilizing large, crowdsourced corpora for training value models, these have often had less-than-desirable performance in practice. 
%The complexity and subtlety of human values make achieving value alignment a challenging task in machine learning models. 
We propose that the solution to this is a high-quality, well-curated dataset designed specifically for conveying value information. 
To that end, we propose the \GG{} Story Corpus. 
This corpus contains a set of three well-curated datasets composed of images and texts to aid in practical value alignment. 
% is essential for facilitating value alignment in these models. In this work, we curated three datasets comprising both images and texts to aid in this effort. 

We also present the performance of baseline models on a set of example tasks. These baselines show that the \GG{} Story Corpus can be used to perform value alignment tasks. It also shows that there is still room for improvement in using the dataset for these tasks, which we hope will encourage researchers to further investigate how this type of data can be used to augment current value alignment systems. 

\section*{Acknowledgements}
We would like to thank Highlights Magazine for allowing us to use and release the Goofus and Gallant dataset. 
%The performance of our baseline models demonstrates the effectiveness of our dataset for value alignment. The high accuracy in identifying normative actions from both images and text indicates that the Goofus and Gallant comic strips contain distinguishable knowledge of normative and non-normative actions. 

%While the GnG text-only normative dataset includes all texts from the comic strips, we could not include all images in the current version due to image quality and technical challenges in automatic extraction. We are continuing to extract more images and collect scene descriptions for these images using human annotators which will be released later as part of our ongoing effort to improve this dataset.

% \section{Conclusion}

% \section*{References}

\bibliography{conference_101719}

%\bibliography{icmla_latex}
\bibliographystyle{IEEEtran}
% \begin{thebibliography}{00}
% \bibitem{b1} G. Eason, B. Noble, and I. N. Sneddon, ``On certain integrals of Lipschitz-Hankel type involving products of Bessel functions,'' Phil. Trans. Roy. Soc. London, vol. A247, pp. 529--551, April 1955.
% \bibitem{b2} J. Clerk Maxwell, A Treatise on Electricity and Magnetism, 3rd ed., vol. 2. Oxford: Clarendon, 1892, pp.68--73.
% \bibitem{b3} I. S. Jacobs and C. P. Bean, ``Fine particles, thin films and exchange anisotropy,'' in Magnetism, vol. III, G. T. Rado and H. Suhl, Eds. New York: Academic, 1963, pp. 271--350.
% \bibitem{b4} K. Elissa, ``Title of paper if known,'' unpublished.
% \bibitem{b5} R. Nicole, ``Title of paper with only first word capitalized,'' J. Name Stand. Abbrev., in press.
% \bibitem{b6} Y. Yorozu, M. Hirano, K. Oka, and Y. Tagawa, ``Electron spectroscopy studies on magneto-optical media and plastic substrate interface,'' IEEE Transl. J. Magn. Japan, vol. 2, pp. 740--741, August 1987 [Digests 9th Annual Conf. Magnetics Japan, p. 301, 1982].
% \bibitem{b7} M. Young, The Technical Writer's Handbook. Mill Valley, CA: University Science, 1989.
% \end{thebibliography}
% \vspace{12pt}
% \color{red}
% IEEE conference templates contain guidance text for composing and formatting conference papers. Please ensure that all template text is removed from your conference paper prior to submission to the conference. Failure to remove the template text from your paper may result in your paper not being published.

\end{document}